\documentclass{article}
\usepackage{amsmath,amssymb}
\usepackage[dvips]{graphicx}   
\usepackage{verbatim}   
\usepackage{color}      
\usepackage{subfigure}  
\usepackage{epsfig}
\usepackage{float}
\usepackage{psfrag}
\usepackage{mathrsfs}
\usepackage{setspace}
\usepackage{footnote}
\usepackage{cite}
\usepackage{hyperref}
\usepackage{authblk}

\newcommand{\bi}{\begin{itemize}}
\newcommand{\ei}{\end{itemize}}
\newcommand{\beq}{\begin{equation}}
\newcommand{\eeq}{\end{equation}}
\newcommand{\bqn}{\begin{eqnarray*}}
\newcommand{\eqn}{\end{eqnarray*}}
\newcommand{\ba}{\begin{array}}
\newcommand{\ea}{\end{array}}
\newcommand{\bs}{\begin{small}}
\newcommand{\es}{\end{small}}
\newcommand{\nn}{\nonumber}

\title{Design and Evaluation of a Tutor Platform for Personalized Vocabulary Learning}
\date{}
\begin{document}
\author[1]{Ravi Kokku}
\author[1]{Aditya Vempaty}
\author[1]{Tamer Abuelsaad}
\author[1]{Prasenjit Dey}
\author[1]{Tammy Humphrey}
\author[2]{Akimi Gibson}
\author[2]{Jennifer Kotler}
\affil[1]{IBM TJ Watson Research Center}
\affil[2]{Sesame Workshop}

\maketitle
\begin{abstract}

This paper presents our experiences in designing, implementing, and piloting an intelligent vocabulary learning tutor. The design builds on several intelligent tutoring design concepts, including graph-based knowledge representation, learner modeling, and adaptive learning content and assessment exposition. Specifically, we design a novel phased learner model approach to enable systematic exposure to words during vocabulary instruction. 
We also built an example application over the tutor platform that uses a learning activity involving videos and an assessment activity involving word to picture/image association. More importantly, the tutor adapts to the significant variation in children's knowledge at the beginning of kindergarten, and evolves the application at the speed of each individual learner. A pilot study with 180 kindergarten learners allowed the tutor to collect various kinds of activity information suitable for insights and interventions both at an individual- and class-level. The effort also demonstrates that we can do A/B testing for a variety of hypotheses at scale with such a framework.



\end{abstract}

\section{Introduction}

The problem of vocabulary gap among students in the early years of school, and the resulting impact on school success have received significant attention in the past ~\cite{BeckMK2013,Biemiller2010,Biemiller2011,Biemiller2015,HartR2003,FarkasB2004}. Early introduction of vocabulary through either direct or indirect instruction helps children learn to read well and forms a strong foundation for literacy, which in turn helps children in accelerated reading to learn. Importantly, while reading new texts, children tend to connect the words they are familiar with to the words exposed in the texts; hence, greater and diverse vocabulary leads to better comprehension of the texts being read.

Given the enormity of vocabulary in English language (and most languages in general), new word acquisition is an ongoing process for many years, and sometimes is even life-long for many people. However, the highest rate of vocabulary development happens in the early years, and teachers in elementary schools focus (often in their own ways, since no universally standardized word lists or procedures exist) a non-trivial amount of time in introducing words to children through both direct and implicit instruction. Implicit or indirect instruction refers to word exposure through reading a variety of leveled texts with the expectation that children infer meanings of unfamiliar words through context~\cite{NeumanW2015}.

Sustaining either direct or implicit instruction is a non-trivial challenge for both teachers and students, and depends on a number of factors including the contexts children are exposed to, their home environments, their interest in continued and varied reading, and the quality and interest of teachers at a school \cite{anderson1984schema,boulware2007instruction,beck2013bringing,vygotsky1978interaction}. Appropriate real-time assessment of the learner understanding is also critical to obtain the necessary feedback on instructions \cite{brown2005automatic,hoshino2005real,lin2007automatic,mostow2012generating}.

To this end, with several recent advancements in artificial intelligence (AI) technologies, we set out to answer the following two questions: (1) how can these advancements be helpful in supporting educators with early vocabulary instruction? And, (2) what specific learning science strategies can be implemented by AI systems at scale and help accelerate vocabulary learning? In this paper, we discuss our experience in tackling this problem by building a layered tutor architecture that enables easily extensible adaptive vocabulary instruction. The extensibility ensures that vocabulary instruction can continue to happen across multiple years through a variety of learning experiences including mobile applications, videos, toys, activities involving tangible interfaces, etc. At the heart of the tutor is a unified learner model that helps ensure that content and assessments are exposed to each individual child in a systematic manner for maximal efficacy of vocabulary acquisition and retention. 

In the rest of the paper, we describe a subset of learning science principles and best practices that we build our solution on, and discuss the design, implementation, and evaluation of our solution in kindergarten classrooms.



\section{Some Best Practices from Learning Sciences}
\label{sec:best-prac}

Christ and Wang~\cite{ChristW2010} examined a number of past efforts on vocabulary instruction in early childhood. Three major approaches emerge across various successful studies: 
\begin{itemize}
\item Early and contextual exposure to advanced words: Words generally known to fewer children engender curiosity in children, thereby leading to a positive effect on continued learning of advanced words. A number of efforts have created lists and categories of words that are amenable to more effective instruction at different stages of learning and in different contexts. These lists form a knowledge base, on top of which customization can be done based on the learning context. 
\item Direct instruction and repetition in a variety of contexts: The more likely children encounter a word, the more likely they understand its meaning~\cite{BeckMK2013}; 
many scholars believe that children need multiple exposures to make words sticky~\cite{biemiller2006effective}. Hence, along with direct instruction of words (especially before reading a story or a rhyme or watching a video), creating the right opportunities (including conversations, videos, rhymes, etc.) for reinforcing the recently learned words could prove quite effective~\cite{horst2011get}.
\item Teacher conversations in meaningful contexts: A critical aspect of instruction in school is the inclusion of teacher talk and conversations using the specific words in contexts~\cite{ChristW2010}.
Beck et al.~\cite{BeckMK2013} 
show that teacher-child conversations provide opportunities for children to demonstrate comprehension. Once deeper insights of which words students are generally struggling with are known, and which words the class in general is already comfortable with, teacher-child conversations could be lot more contextualized and personalized. 
\end{itemize}

Additionally, carefully instrumenting learning and assessment activities and gathering data can enable evidence-driven insights of children's understanding of words systematically. 
We now describe the design of a vocabulary tutor that attempts to realize these best practices at scale using a systematic combination of technologies. 


\section{The Vocabulary Tutor Architecture}
\label{sec:tutor-arch}

\begin{figure}[t]
\includegraphics[width=4.8in]{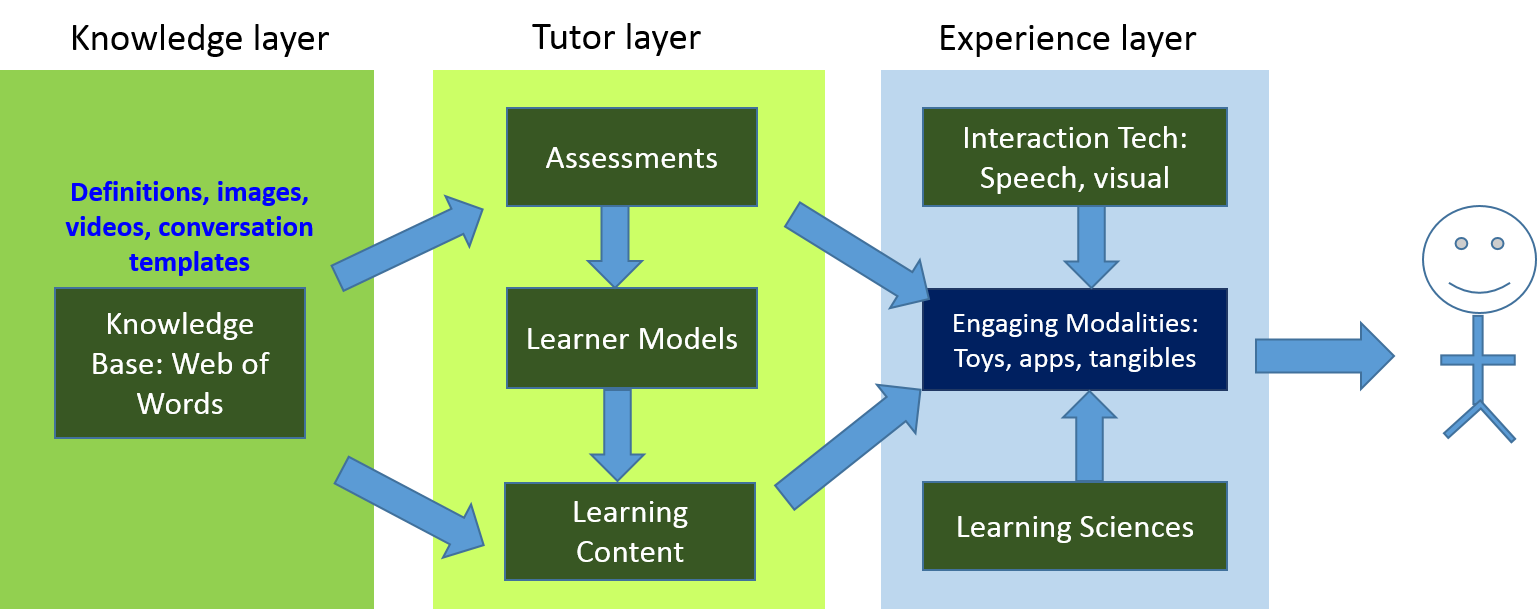}
\caption{The Vocabulary Tutor Layers.}
\end{figure}

Intelligent tutoring systems (ITS) is a well-studied subject for a variety of domains and ages \cite{Bloom1984,CohenKK1982,CummingM2000,Woolf2008,Prentzas2013}. Our vocabulary tutor builds on this literature and is designed as three layers: a knowledge layer, a tutor layer, and an experience layer. The knowledge layer represents domain information in a manner that is easily consumable by the tutor layer for enabling intelligent experiences to the learner. Building on the AI research of knowledge representation and reasoning that has shown use of conceptual networks as powerful knowledge representations \cite{Sowa1983,CheinM2008}, our knowledge layer is represented as a web of words and concepts: each node in the word web represents a word or a concept, and each link between a pair of nodes represents one or more relationships between them. A number of word and common sense databases (e.g. WordNet \cite{WordNet}, ConceptNet \cite{speer2017conceptnet}, OpenCyc \cite{lenat1993building}), along with hand-curated information by subject matter experts can form the basis of this knowledge layer. This layer represents content as an easily query-able knowledge base that will also index into learning and assessment content including images and learning videos related to the word concept. The layer can be easily extended to other forms of learning content such as stories and rhymes. For instance, in the context of vocabulary learning, a word such as \emph{Habitat} will have definition of the word, images showing the concept of the word Habitat in different settings, videos explaining the concept, and usage of the word in different sentences that are used for conversations. 

The knowledge base, represented as a graph data structure including nodes and links, also helps derive assessments automatically (which can be filtered later by a subject matter expert) using inter-word relationships through automatic graph-traversal techniques. One such assessment type we generate from the knowledge base is a set of picture-based multiple choice questions. For each word, the process generates a number of [{\em Which picture represents the word $<$WORD$>$?}] questions, each with one correct image and two distractor images. Our approach leverages the observation that creating new and customized assessments by a human is in general harder than verifying the auto-created ones. Although, this verification process might be skipped in general, due to the nature of early-childhood learners,  verification of age-appropriateness is necessary in our process.

Exposing a variety of these assessments repeatedly over a period of time in different gamified contexts can help establish more confidently the child's understanding of a word or a concept, as testable by the particular assessment type. Observe that different assessment types test different levels and dimensions of understanding of a word, and hence a combination of these assessment types are needed in a vocabulary tutor for developing confidence in the tutor that a learner \emph{knows} a particular word. For instance, the ability of a learner to associate a word with a picture does not necessarily mean that the learner can use the word in a sentence. 

In the process of enabling the exposure of different assessments to children, the tutoring platform builds learner models that are transferable across multiple activities in the experience layer. For instance, the fact that a child has learned a word \emph{Desert} in one mobile application can be used by another application or a toy to expose \emph{Desert} in the context of a game or a conversation or an activity. Learner models \cite{DesmaraisB2012,Brusilovskiy1994} form the basis for personalization and adaptation as they use learner's past behavior data to \emph{learn} and \emph{predict} their future behavior. Learner models (and their learning algorithms) can vary in complexity depending on model representation and amount of data available, ranging from simple graphical model-based such as Bayesian Knowledge Tracing (BKT) \cite{CorbettA1994}, to the more recent deep learning-based Deep Knowledge Tracing (DKT) \cite{Piech2015}.
To exemplify the scalability of our tutor, for our purposes, learner model is essentially a set of learner scores for each word in different dimensions. For instance, we currently maintain learner scores in four dimensions as a confidence in the learner's proficiency in each of the four dimensions: listening, reading, speaking, and writing. Each application built over the tutor and assessing a learner updates the scores in one or more of the dimensions. 

Using the learner models, the tutor layer enables personalization of assessments and learning content. Both the learning and assessment content can be presented to learners in a variety of experience modalities, including speech, vision, and touch interactions. Due to the rapid pace at which AI-technologies are evolving, we envision use of several other experience modalities such as AR/VR, and hence our tutoring platform is developed to accommodate variety of experiences. The design of these experiences could be guided by learning science principles, as chosen by developers building the applications in the experience layer. 

This combination of knowledge representation and retrieval, automatic assessments, and personalization enable the tutoring platform to support continuously evolving learning activities as a learner's knowledge increases with each exposure. We now describe the phased learner model design and implementation in more detail.


\subsection{Phased Learner Model}
\label{sec:PLM}
\begin{figure}[t]
\includegraphics[width=4.8in]{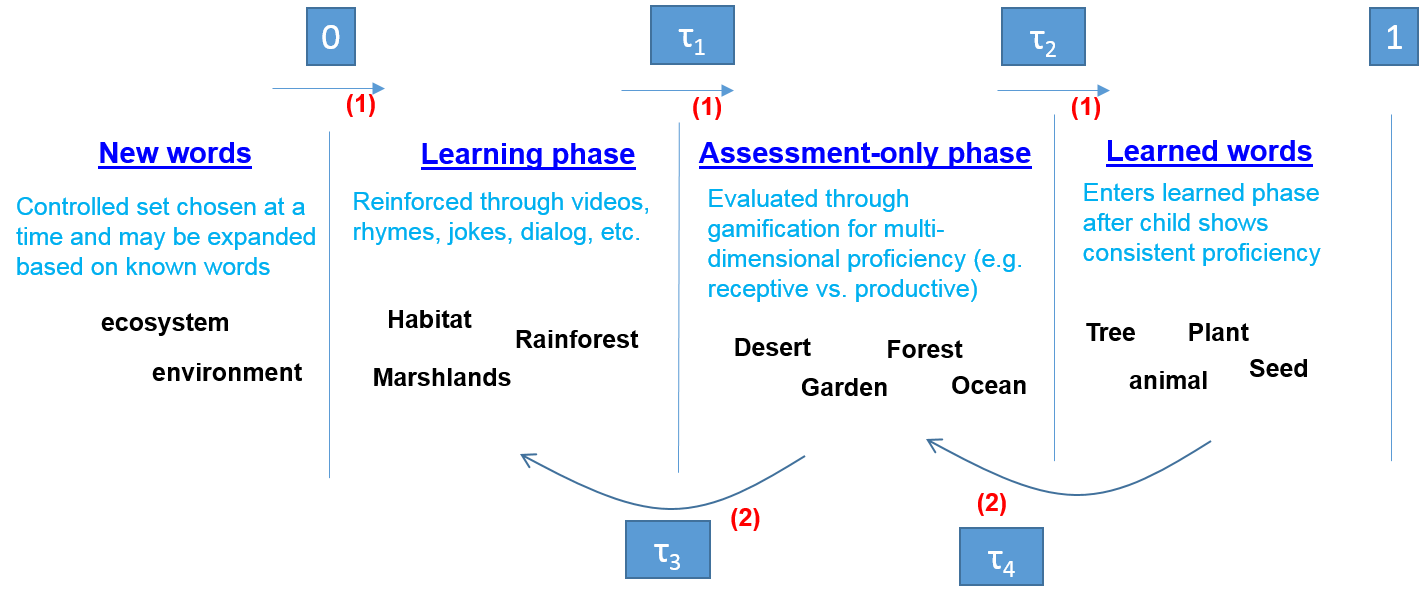}
\caption{Phased Learner Model showing words in different phases based on learner score, and phase transition thresholds.}
\label{fig:PLM}
\end{figure}

The goal of the phased learner model (Fig.~\ref{fig:PLM}) is to enable systematic exposition of words, and enable repetition on a controlled set of words at any point of time. At a logical level, we consider 4 phases in which a word can be: New word phase (also referred to as Parked phase), Learning phase, Assessment-only phase, and Learned phase. Each word has a learner score between 0 and 1, and different thresholds on the learner score determine which phase a word is in for that learner: e.g. when the score crosses the threshold $\tau_1$, the words move from Learning to Assessment-only phase. The learner score starts at zero, and builds up towards 1 as assessments are exposed and individual performance scores are received. In our implementation, we use an EWMA (Exponentially weighted moving average) based update of learner score for each new performance score: $$l_{t+1} = \alpha l_t + (1- \alpha)s_{t+1},$$ where $l_{t+1}$ and $l_{t}$ are learner scores at time $t$ and $t+1$ respectively, $s_{t+1}$ is the score of the $(t+1)$-th assessment, and $\alpha$ is the EWMA parameter.

Words unexposed by the tutor to a learner (and hence the tutor lacks confidence of the learner's proficiency of the word) are in the New words phase. The {\em working set} of words at any point of time includes words in the two phases: the Learning phase and the Assessments-only phase. Words in the working set are used at higher priority by applications in the experience layer to enable different learning and assessment experiences. As assessments result in increasing learner score, words transition from Learning phase into Assessment-only phase, where no new learning activities for those words are exposed to a learner. With further assessment, the learner score builds up and the words move into the learned phase. The arrows numbered (1) indicate this flow of words from one phase to another as the scores increase from 0 to 1. 

Note that at each exposure of the assessment, the instantaneous score $s_t$ can be zero, and hence words could move back into previous phases (as determined by thresholds $\tau_3$ and $\tau_4$. To avoid oscillation of words across phases, $\tau_3<\tau_1$ and $\tau_4<\tau_2$). This is indicated by the arrows numbered (2). 
New words are introduced based on related-word expansion: we add new words based on words in Learned and Assessment-only phases to maximize the chance that related words are learned in clusters. New words could also be introduced into the system by a teacher-facing application to facilitate curriculum-aligned instruction. Words are moved from New words to Learning phase incrementally as existing working set of words decreases with child's mastery of words.

\begin{figure}[t]
\includegraphics[width=4.8in]{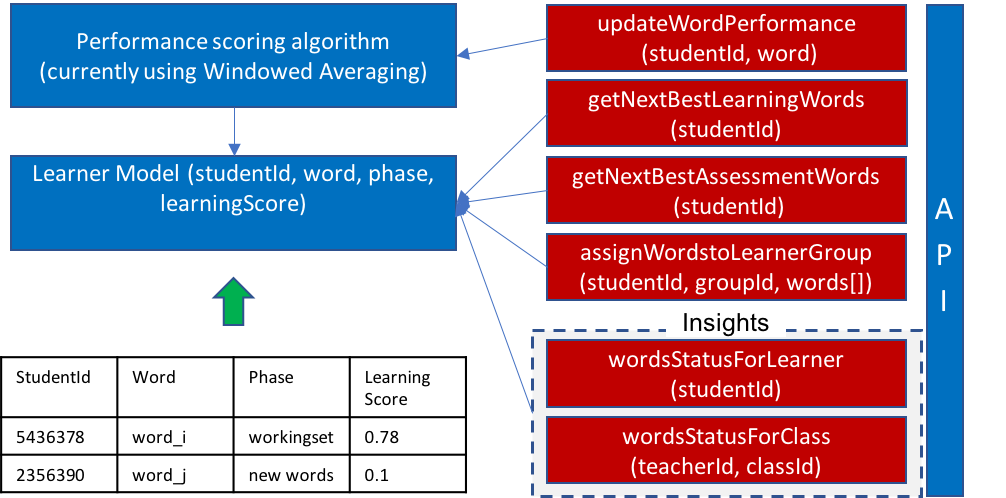}
\caption{Tutor Layer APIs.}
\label{fig:tutor-api}
\end{figure}

Figure~\ref{fig:tutor-api} shows a high level view of the APIs the tutor layer exposes to enable applications in the experience layer. The APIs {\sf getNextLearningWords} and {\sf getNextAssessmentWords} provide a simple interface to applications to access the working set for each learner to expose targeted learning and assessment activities respectively. The {\sf getNextAssessmentWords} function creates a blend of Assessment-only words, words in the Learning phase, and words in the Learned phase to ensure that data samples are continuously collected on words to maintain the most recent confidence of a learner's proficiency on each word. Each application can update the performance for words assessed using the {\sf updateWordPerformance} API. For A/B testing, one could assign specific words to learner groups using {\sf assignWordstoLearnerGroup}. Finally, {\sf wordStatusForLearner} and {\sf wordStatusForClass} are examples of many insights we derive with the data for teacher and administrator dashboards.

\section{Experimental Evaluation}
\label{sec:evaluation}

In this section, we describe the results of evaluation of the tutor in a pilot study conducted in real classroom settings. The pilot study was conducted during a 9-week period in 8 different classrooms spread across 4 schools in Georgia between October and December 2017. In what follows, we first discuss the application we built over the platform that learners use, and then discuss the evaluation methodology and results.

\subsection{Application} 

We built an iPad application that leverages the knowledge and tutor layers of the platform, and enables building learner models for vocabulary learners. The application exposes word-centric videos for learning, which expose words in various contexts and usage scenarios. The application then tests receptive vocabulary via word-to-picture association (i.e., picture based multiple choice questions). Both learning videos and assessments shown to a given learner is driven by his/her current working set. Only words in the learning phase are used to show targeted videos to the learner, and words in the learning and assessment-only phases are mainly used for assessments. Words in the Learned phase are also randomly chosen to test for continued understanding. If the learner does forget the word and gives a wrong answer, the learning score reduces, and the word moves back into the working set.

To ensure that enough data is collected, a particular ratio of learning-to-testing is maintained by the application by enforcing testing activity whenever the relative learning activity crosses a threshold. As discussed in Sec.~\ref{sec:tutor-arch}, this particular application, built in the experience layer, is updating the learner score in listening comprehension dimension, and was presented to the learner using predominantly vision and touch interactions.


The application focused on $K=40$ advanced vocabulary words (including a combination of Tier-2 and Tier-3 words) that have been picked by subject matter experts as target words to be taught to the learners.

\subsection{Experimental Setup}
The pilot study served two purposes. Firstly, it allowed the tutor to be exercised across a diverse set of early kindergarten learners across multiple classrooms. Secondly, and more importantly, it allowed us to demonstrate that the tutor mechanisms we developed can be used to conduct various automated A/B tests atop, and have fine-grained online control on the test process.

To demonstrate the second functionality, while collecting data for {\em every} word, {\em and} including {\em every} learner in the experimental group, we designed the A/B testing methodology in the following manner: the classrooms were randomly split into two groups (Group A and Group B), with all learners within a class belonging to the same group. Similarly all vocabulary words were divided into two word sets (Word Set $X$ and Word Set $Y$). For Group A learners, learning activity was available only on word set $X$, while they were assessed on both word sets $X$ and $Y$. Similarly, Group B learners had learning activity only for word set $Y$, while they were assessed on both word sets $X$ and $Y$. In this way, for words in word set $X$, Group A learners acted as the experimental group and Group B learners were the control group. And for words in word set $Y$, Group B learners acted as the experimental group and Group A learners were the control group. This ensured that every learner belonged to the experimental group and took part in learning activities for some subset of vocabulary words. 

Note that this is just an example, other forms of A/B grouping can be done similarly with online division of learners and words into groups as the system is functional in real settings. The tutor supports associating the words and learners to different groups dynamically online, and automatically adapts the working set. Table~\ref{table:AB} summarizes the A/B testing methodology that we conducted as a part of the pilot.

\begin{table}
\begin{center}
\caption{A/B Testing Methodology}
\begin{tabular}{|c||c|c|}
\hline
 & Group A & Group B\\
\hline
\hline
Number of classes &4 & 4\\
\hline
Number of learners &91 & 90\\
\hline
Learning activity & Word set $X$ & Word set $Y$\\
\hline
Assessment activity & Word set $X\cup Y$ & Word set $X\cup Y$\\
\hline
\end{tabular}
\end{center}
\label{table:AB}
\end{table}

\subsection{Metrics of Evaluation}

To demonstrate the efficacy of the tutor, we divide the metrics into two categories, which are aligned with the two purposes the study serves. For the tutor's personalization based on learner diversity, we show different snapshots of the tutor's working set and learned set across learners. We also show that this data when exposed to the teachers can provide valuable insights into which words need manual intervention at higher priority at a class level, and which words need personalized intervention.

To demonstrate how A/B tests can be used for deriving insights across different tutor activities, we develop the following comparison methodology.
%
Let there be two sets of assessment response data ($G_1$ and $G_2$) each consisting of multiple assessment response vectors ($N_1$ vectors in $G_1$ and $N_2$ in $G_2$). For vector $l$ where $l=1,\ldots,N_1$, let $\mathbf{a}_l\in\{0,1\}^{t_l}$ be the $t_l$-length bit sequence representing the assessment responses.\footnote{$a_l$=1 if the assessment response is correct and $0$ otherwise.} 
 
The goal is to determine if the data vectors in $G_2$ are \emph{significantly better} than the data vectors in $G_1$. In other words, the distribution of data in $G_2$ is \emph{significantly greater} than the distribution of data in $G_1$. This can be modeled as the following one-sided hypothesis test.
\begin{eqnarray}
H_0: f_\mathbf{G_1}=f_\mathbf{G_2}\nn\\
H_1: f_\mathbf{G_1}<f_\mathbf{G_2}
\label{eq:hyp_gen}
\end{eqnarray}
where $f_\cdot$ represents the probability distribution, and $<$ is defined in the sense that the data values in $G_1$ tend to be \emph{larger} than those in $G_2$, or in other words the cumulative distribution function of $G_1$ tends to be smaller than that of $G_2$. These kind of statistical hypothesis tests help us evaluate the success or failure of tutoring interventions. 

Note that in such a general formulation, each data vector $\mathbf{a}_l$ can be of different length. Hence, determining the distribution of the vectors within the group is not straightforward. To address this concern, we make the following simplification: for each assessment response sequence $\mathbf{a}_l$, the assessment responses can be considered to be independently and identically distributed binary random variables and hence, the mean $p_l$ of $\mathbf{a}_l$ is the sufficient statistic to represent $\mathbf{a}_l$. Note that this is a valid assumption especially when the underlying unknown distribution is Bernoulli distributed. 
Based on this discussion, \eqref{eq:hyp_gen} simplifies to testing if data $\mathbf{p}_1=[p_1,\ldots,p_{N_1}]$ and data $\mathbf{p}_2=[p_1,\ldots,p_{N_2}]$ come from significantly different distributions. In other words
\begin{eqnarray}
H_0: f_{\mathbf{p}_1}=f_{\mathbf{p}_2}\nn\\
H_1: f_{\mathbf{p}_1}<f_{\mathbf{p}_2}.
\label{eq:hyp_pq}
\end{eqnarray}
 
This test can be performed using the typical one-sided two-sample hypothesis tests such as t-test of means \cite{ChakravartyRL1967} or one-sided two-sample Kolmogorov-Smirnov (KS) test \cite{ChakravartyRL1967}. 
Note that, the assessment sequence $\mathbf{a}_l$ is abstracted using mean $p_l$. To ensure statistical significance, we can only consider the assessment vectors in $G_1$ and $G_2$ for which $t_l\geq\tau$. Also, while comparing distributions of $\mathbf{p}_1$ and $\mathbf{p}_2$, we need to ensure there is a lower limit $\eta$ on the size of $G_1$ and $G_2$. In other words, the analysis is performed only when $N_1\geq\eta$ and $N_2\geq\eta$.

\subsection{Results}

We now present the results obtained from the pilot study. For the study, the parameter values of the tutoring system used are summarized in Table~\ref{table:parameters}.


\begin{table}
\begin{center}
\caption{\label{table:parameters}Parameter values for experimental evaluation.}
\begin{tabular}{|l|c|}
\hline
Parameter & Value\\
\hline
\hline
EWMA parameter ($\alpha$) & 0.8\\
\hline
Threshold to transition into Learned Phase ($\tau_1=\tau_2$) & 0.86\\
\hline
Threshold to transition out of Learned Phase ($\tau_3=\tau_4$) & 0.56\\
\hline
$\tau$ & 3\\
\hline
$\eta$ & 10\\
\hline
\end{tabular}
\newline
\newline $\tau$ - minimum number of assessment responses needed
\newline $\eta$ - minimum number of learners needed in each of the groups
\end{center}
\end{table}

\subsubsection{Adaptation and Personalization}

One of the key requirements of an intelligent tutor is the power to personalize and adapt content based on the learner behavior. In other words, the tutor must \emph{learn} about the learner, build a model on their understanding of material, and serve content as per their learning pace and interests. Our AI-powered tutored ensures this personalization and adaptation using the phase learner model described in Sec.~\ref{sec:PLM}. 

\begin{figure}[t]
\centering
\includegraphics[width = 3.2in, height=!]{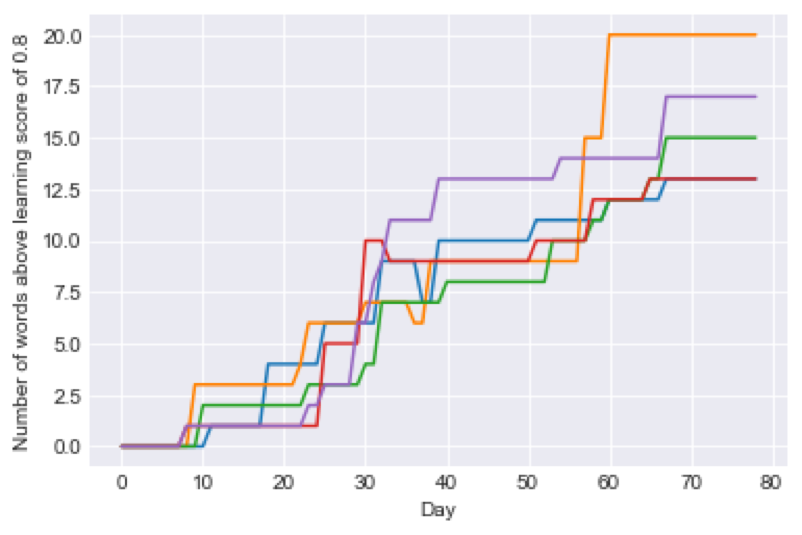}
\caption{Learning rates for different users. }
\label{fig:learningrate}
\end{figure}

Fig.~\ref{fig:learningrate} shows a count of words with a high learning score over time for $5$ learners. The plot shows that while all learners started with the same unknown confidence and low number of words with a high learning score, the pace of learning is different for every learner. This observation shows quantitatively that tutor-enabled learning activities in classrooms can enable the much needed personalization at scale, by complementing the teachers, and letting them only focus on words or concepts that the class as a whole is unable to grasp. 

\begin{figure}[t]
\centering
\includegraphics[width = 4.5in, height=!]{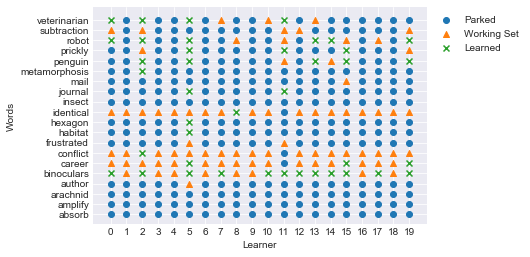}
\caption{Snapshot of learner model phases for randomly picked learners.}
\label{fig:phases_snapshot}
\end{figure}

The adaptation and personalization of the tutor is more striking in Fig.~\ref{fig:phases_snapshot} where the learner model phases of different words is presented on day $50$ of the pilot for $20$ randomly chosen learners. The \emph{blue circle} represents the \emph{Parked} phase, the \emph{orange triangle} represents the \emph{Working Set}, and the \emph{green cross} represents the \emph{Learned} phase. As the figure shows, the system personalizes and adapts the learning material presented to the learner based on their phased learner model, which is learnt and updated using the assessment responses.

\subsubsection{A/B Testing Result}
We now move to a simple question using the results obtained from groups A and B: Is the assessment performance of learners higher for learners in the experimental group than the control group? This question is answered individually for every vocabulary word concept. To answer this question, we set $G_1$ to be the set of assessment responses of all learners that belong to the control group for the particular vocabulary word (Group B learners for words in word set $X$, and vice versa), and $G_2$ to be the set of assessment responses of all learners that belong to the experimental group for the particular vocabulary word (Group A learners for words in word set $X$, and vice versa). 

\begin{figure}[t]
\centering
\includegraphics[width = 4in, height=!]{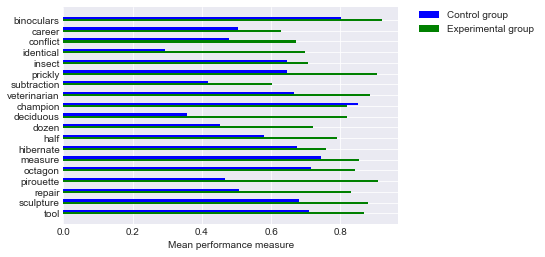}
\caption{Performance comparison between control and experimental groups across multiple learning concepts.}
\label{fig:hyp2}
\end{figure}

Fig.~\ref{fig:hyp2} compares the average performance of learners in experimental and control groups. As the figure shows, the performance on word-image association test increases among experimental group in almost all the concepts (except \textit{champion} for which it remains almost same). This increase was statistically significant ($p<0.1$) for $16$ of the $19$ word concepts. Note that this observation by no means is conclusive on the fact that learners really know the word to an extent that they can use the words in practice, can define the words when asked, can associate the words in all contexts they occur; the observation only demonstrates the difference in word-picture association in experimental and control groups, and mainly demonstrates the tutor's ability to conduct such comparison tests at scale with just online mapping of words and learners to specific experimental groups. This also presents the ability to provide insights to the teacher on the words in the curriculum that are well-known by the entire class and hence are of least priority for manual intervention. For example, in Fig.~\ref{fig:hyp2}, the words such as \emph{binoculars}, \emph{champion}, and \emph{measure}, have a high performance among the control group itself implying a generally high understanding of the word without intervention. Therefore, words such as \emph{deciduous}, \emph{identical}, and \emph{subtraction}, can be prioritized for manual intervention.

\section{Paper and Pencil Tests}
To assess whether the learning that was captured by the tutor was also evident in the more common format of classroom assessments, we asked the teachers in the 8 classrooms to select 5 children in their class to whom they would administer paper-pencil baseline and endline tests. These tests were constructed such that for each word, the children were asked an open ended question: such as ``what is an octagon" followed by a multiple choice test with 4 choices such as ``which one of these is an octagon." There were 31 words in the test booklets and teachers were free to administer the tests in one or more sessions both at baseline and endline. This method allowed us to measure expressive language (in this case children's ability to describe each word) and receptive language (in this case ability to identify words through pictures). There were 40 children who were pre-tested and all children were post-tested but several had missing data on various words. The teachers did not know which words the tutor had assigned children to learn, thus making the teacher relatively blind to the intervention. We suggested that they choose any 5 children in their class. The teachers, in the orientation session, suggested that they choose a mix of children who learn quickly and those who struggle.

\subsection{Quantitative Analysis}
In order to assess the relationship between the tutor's assessment and the teacher administered test, we examined the difference in the teacher administered test from pre-to-post between children who were trained on the words by the tutor and those who were not. There were only 6 words of the 31 that 10 or more children (of the 40 selected) were trained on. These were: octagon, sculpture, veterinarian, identical, subtraction, and deciduous. While we have also calculated the percent correct between those who were trained versus those who were not trained for words with fewer than 10 children, quantitatively they cannot be measured with confidence. The percent correct to items in the teacher assessment are shown in Table~\ref{table:paper-pencil} below. 

\begin{table}[htp]
\caption{Percent correct to multiple choice items in teacher assessment; comparison on whether children were trained by the tutor}
\begin{center}
\begin{tabular}{*7c}
\hline
 &  \multicolumn{3}{c}{UNTRAINED} & \multicolumn{3}{c}{TRAINED}\\
 \hline
 Word & $N$ & Baseline & Endline & $N$ & Baseline & Endline\\
 \hline
\hline
 octagon &	25	& 84\%	&	80\%		&	15	& 80\%	&	93\%\\
 \hline
 sculpture	& 28	& 79\%	&	93\%		&	12	& 58\%	&	100\%\\
 \hline
 veterinarian &	30 &	63\%	&	70\%		&	10	& 50\%	&	100\%\\
 \hline
 identical	& 28	& 18\%	&	29\%		&	12	& 33\%	&	92\%\\
 \hline
 subtraction &	28	& 29\%	&	21\%		&	11	& 18\%	&	18\%\\
 \hline
 deciduous &	26	& 12\%	&	19\%		&	13	 & 0\%	&	92\%
\end{tabular}
\end{center}
\label{table:paper-pencil}
\end{table}%

For all but ``subtraction", the children in the trained group had considerably higher scores at endline than the children who were not trained on a particular word; even though in most cases, the trained children had lower scores at baseline than the untrained group. This suggests that the tutor's assessment of children's learning as indicated by multiple choice tests were \emph{accurate} depictions of how children would score on multiple choice tests in the teacher administered test (which is how common assessments are usually administered). 

\subsection{Qualitative Analysis}
Children in the trained group also exhibited richer and more accurate descriptions of words in their free response at endline compared to baseline.  Some examples are shown in Table~\ref{table:qual} below.

\begin{table}[htp]
\caption{Example responses to open-ended questions}
\begin{center}
\begin{tabular}{|c|p{2in}|p{2in}|}
\hline
Word & Baseline & Endline\\
\hline
\hline
Octagon & ``I don't know" & ``An 8 sided shape"\\
\hline
Sculpture & ``You like it. It's a painting. You let it sit a moment to dry"	 & ``Made of metal, rock, marble and clay"\\
\hline
Veterinarian & ``When something sticks onto something" & ``Taking care of a pet"\\
\hline
Identical & ``Something that looks the same but it is not the same" & ``People and things that look exactly the same"\\
\hline
Subtraction & ``Stick metal to metal" & ``Take away"\\
\hline
Deciduous & ``When something is rotten" & ``When a tree loses its leaves once a year"\\
\hline
\end{tabular}
\end{center}
\label{table:qual}
\end{table}%

\subsection{Summary}
Given that the teacher assessment captured children's learning in a similar fashion to the tutor (at least on the 6 words that there was enough comparable data to assess), we can conclude that the algorithm used to designed the tutor's assessment has at least some predictive validity. Going forward, it will be important to power the sample such that statistical analyses can be conducted to truly understand the relationship between the tutor and the assessments that school systems use to measure progress in children.

\section{Discussion and Conclusion}
\label{sec:concl}

In this paper, we describe the design and implementation of a vocabulary tutor platform that builds on ITS concepts and enables building a variety of learning experiences over a phased learner model. We described and presented results from a pilot study conducted in real classrooms. Using the A/B testing implemented on the platform, we presented evidence of learning by learners upon using the tutor. Quantitative and qualitative analysis of paper-and-pencil tests conducted on randomly picked students also confirmed our results that learners learnt new vocabulary words using the tutor platform.

The specific learning and assessment activities we developed are only exemplars and limited in the ways in which words are exposed and tested for understanding. However, the effort does reveal the advantage of developing a unified tutor framework for a variety of applications enabling learning experiences that can personalize vocabulary learning at scale, and target addressing the 30 million word gap problem. More importantly, even though explicit vocabulary instruction is taken up during early school years (e.g. kindergarten to 4th grade), retaining acquired vocabulary requires continued repetition of exposure and usage in different meaningful contexts. We envision that our tutoring solution provides such a life-long word concept learning platform as newer applications are developed and newer domains are explored.



\bibliographystyle{splncs03}
\bibliography{vempaty_lib}
\end{document}